\documentclass[11pt]{article}

\usepackage[final]{acl}

\usepackage{amsmath}  
\usepackage{amssymb}
\usepackage{times}
\usepackage{latexsym}

\usepackage[T1]{fontenc}

\usepackage[utf8]{inputenc}

\usepackage{microtype}

\usepackage{inconsolata}

\usepackage{graphicx}
\usepackage{stfloats} 
\usepackage{afterpage}
\usepackage[noabbrev]{cleveref}
\usepackage{booktabs}
\usepackage{subfigure}
\usepackage{subcaption}
\usepackage[table]{xcolor}
\usepackage{enumitem}
\usepackage{xcolor}
\usepackage{multirow} 
\usepackage{placeins}
\usepackage{algorithm}
\usepackage{algorithmicx}
\usepackage{algpseudocode}
\usepackage[most]{tcolorbox}
\usepackage{makecell}
\usepackage{float}
\crefname{table}{Table}{Tables}
\crefname{figure}{Figure}{Figures}

\floatname{algorithm}{Algorithm}

\title{MASS: Deep Research for Social Sciences with Memory-Augmented Social Simulation}

\author{
	Yongrui Liu\textsuperscript{1,2}, Deyi Xiong\textsuperscript{2\thanks{Corresponding author.}} \\
	\textsuperscript{1}The International Joint Institute of Tianjin University, Fuzhou,
	Tianjin University,	China \\
	\textsuperscript{2}TJUNLP Lab, School of Computer Science and Technology, Tianjin University, China\\
	\texttt{\{liuyongrui,dyxiong\}@tju.edu.cn}
}

\begin{document}
\maketitle

\begin{abstract}
Deep Research agents powered by Large Language Models (LLMs) have exhibited extraordinary potential in automated paper writing tasks. However, existing systems rely heavily on literature retrieval and synthesis through internet and local knowledge bases, often resulting research in lacking insight and creativity in social science. To address this issue, we propose ``Memory-Augmented Social Simulation (MASS)'', an innovative paradigm that leverages highly realistic and research-oriented social simulations to enhance the creativity and empirical founding of LLMs-generated research. Specifically, MASS integrates three core components: dynamic goal-path planning with multi-level social norm restraint to guide the simulation, a multi‑disciplinary behavior dataset for agent memory cold‑start, and a structured forgetting mechanism inspired by the Ebbinghaus curve. Together, these ensure simulation authenticity and provide a robust empirical foundation for generating innovative scholarly papers. Experimental results demonstrate the effectiveness of our method, showing a $6.81\text{\%}$ improvement in generation overall quality over foundation LLMs and $17.19\text{\%}$ gain in Insight over strong baselines. Dataset and codes will be released.\footnote{\url{https://github.com/tjunlp-lab/MASS_DeepResearch}}
\end{abstract}

\section{Introduction}
Large Language Model agents have achieved breakthrough progress, demonstrating exceptional capabilities in knowledge discovery \cite{liu-etal-2024-lhmke,guo-etal-2024-ctooleval}, information integration \cite{guo2023evaluatinglargelanguagemodels}, and multi-agent collaboration. This potential is gradually permeating academic research, spurring exploration into automated scientific research and data analysis \cite{xu2025comprehensive,zhang2025deep}.

Among them, Deep Research agents have been widely applied in social science, which is an autonomous research agent that executes workflows and generates papers by planning and iterative retrieval, a process dependent on retrieving and synthesizing vast literature. They, however, prove particularly limiting in social science, a field characterized by complexity, practicality and subjectivity, where true insight arises from the interplay between theory and practice \cite{ashworth2021theory,gerxhani2022handbook}. Merely reprocessing existing texts generates theoretical insight and academic creativity that are insufficient to social science research.

To address this challenge, we propose an innovative paradigm for automated social scientific writing, Memory-Augmented Social Simulation (MASS). Departing from the conventional ``retrieve-and-generate'' mechanism of Deep Research Agent, MASS constructs a realistic and research-oriented virtual social simulation environment. Under such setting, LLMs are guided to conduct exploratory and creative social simulation, thereby generating original, empirical “first-hand” data that provides a substantive foundation for subsequent paper writing.

Specifically, the MASS framework consists of four tightly integrated core modules:
1) The \textbf{Task Planning and Deep Reasoning Module}, which employs a divergence Chain-of-Thought (COT) structure to generate diverse research hypotheses and conducts multi-dimensional evaluations to determine the optimal research pathway \cite{zhu2025scaling};
2) The \textbf{COT Execution and Tool Dispatching Module}, which
\afterpage{
	\begin{figure*}[t!] 
		\centering
		\includegraphics[trim=12 11 12 11, clip,width=1.0\textwidth, keepaspectratio]{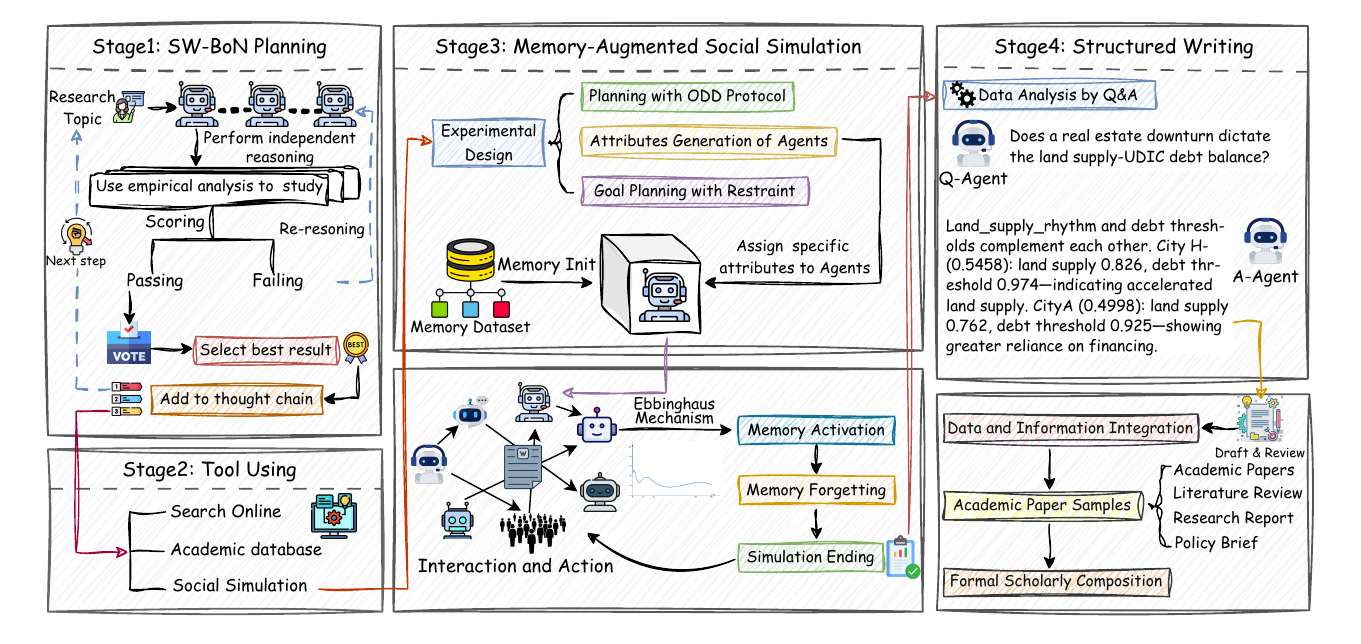}
		\caption{Diagram of the MASS Deep Research framework. It involves four steps: (1) Using divergence COT to plan the process, (2) Calling tools, (3) Agents interact with the memory module, (4) Analyzing retrieved and simulated data, writing with formal style.}
		\label{fig:simulation_framework}
	\end{figure*}
}
translates the chosen pathway into executable steps, including retrieving knowledge via internet access and conducting social simulation experiments to generate original empirical data;
3) The \textbf{Social Simulation Experiment Module}, which represents the core innovation of MASS to standardize experimental design using the ODD protocol \cite{grimm2020odd} and establishes a virtual social environment with dynamic goal planning and social norm constraints. Agents are endowed with a memory system grounded in multidisciplinary knowledge and an Ebbinghaus‑inspired forgetting mechanism, ensuring authentic and reliable simulations.
4) The \textbf{Structured Writing Module}, which adheres to academic templates and genre conventions, formatting and composing the final output to meet scholarly standards \cite{becker2008writing}.

We present a paradigm shift in Deep Research, enhancing the research pipeline with a generative social simulation that produces novel empirical evidence, as illustrated in \cref{fig:simulation_framework}. The main contributions are summarized as follows:
\begin{enumerate}[leftmargin=*, label=\normalfont\centering\textbullet, align=left]
	\item \textbf{The MASS Paradigm}: We propose Memory-Augmented Social Simulation paradigm, a novel framework that advances automated social science research beyond conventional approaches.
	\item \textbf{Dataset \& Memory Mechanism}: We develop a multi-disciplinary social behavior dataset and an Ebbinghaus-inspired memory mechanism, significantly enhancing the authenticity and cognitive continuity of agent behavior.
	\item \textbf{Goal-Planning with Norm Constraints}: We design a dynamic goal-planning scheme under multi-layered social norm restraint, which imbues the core simulation process with a clear research orientation.
\end{enumerate}

\section{Related Work}
	\textbf{Deep Research Agent.} The development of Deep Research agent systems generally progress through several key phases. Early primarily adopt a single-agent architecture \cite{hilton2022webgpt,Yao2022ReAct}. Subsequently, multi-agent architectures emerge to enable specialized division of labor \cite{tang2025autoagent,Zheng:2025emnlp}. Furthermore, hybrid architectures subsequently arise, aiming to balance the flexibility and efficiency \cite{saied_2025_perplexity}. Recent end-to-end paradigms incorporate Reinforcement Learning to enhance reasoning capabilities \cite{TongyiDeepResearch2025,wan-etal-2025}. While existing studies focus on architectural optimization for general research tasks, our work introduces creative design from a specialized simulation paradigm in social sciences, enchancing the innovativeness of research.
	
	\textbf{Agent Behavior and Decision-Making.} Research on agent behavior and decision-making is primarily categorized into three paradigms. The static planning paradigm predefines a complete action sequence. \cite{lei2025rhinoinsight,zhang2023autoparallelizing}. And the dynamic planning operates through a ``perception-decision-action'' cycle for real-time interaction \cite{wu2025autono,Yao2022ReAct}. The dual-system synergistic paradigm further integrates both approaches to balance rapid response and long-term objectives \cite{chen2025mars,li-etal-2025-chatsop}. Building on these paradigms, our MASS framework enhances agent autonomy via multi-level constraint and dynamic goal planning.
	
	\textbf{Social Simulation.}  Existing LLM-based multi-agent social simulation studies advance the field from three key dimensions. First, data injection and fine-tuning techniques improve behavioral authenticity in micro-level simulations \cite{merrill2025pointoforder,PMC7743915}. Next, to lower the technical barrier, large-scale asynchronous parallel social experiment simulation platforms also emerge \cite{gao2023s3,wang2025yulanonesim,piao2025agentsociety}. Meanwhile, mean-field theory is introduced to reduce computational costs in agent-group interactions \cite{mi2025mf}. Compared to these advances, our framework reduces long-context overhead and further enhances simulation authenticity via a dedicated memory augmentation module.

\section{MASS Deep Research}
	The Deep Research framework is grounded in Memory-Augmented Social Simulation (MASS). The framework shifts from the passive integration of information to the active generation of experiential insights, directly facilitating the production of social science research papers.
	
\subsection{Task Planning and Deep Thought}
	In complex reasoning tasks, traditional chain-of-thought approaches, such as COT and its Best-of-N extensions, can enhance single-step logic but often struggle with limited exploration in multi-step, highly divergent tasks \cite{wei2022chain,yao2023tree}. To improve the decision-making exploration capability of Deep Research in task planning, enable stronger divergent thinking in tackling complex tasks, effectively address the ``error accumulation'' problem in multi-step reasoning tasks, and enhance both the coherence and outcome diversity, we propose a Stepwise Best-of-N (\textbf{SW-BoN}) strategy.
	
	 Distinguished from the traditional COT and its extended form COT Best-of-N, the core of our strategy lies in expanding the search space of deep research through stepwise multi-path exploration and real-time optimal selection within each stage. Specifically, we define the SW-BoN as follows. At step $i$ of the task, agents first generate $n$ candidate reasoning nodes, denoted as $\boldsymbol{C}_i = [c_{i1}, c_{i2}, \dots, c_{in}]$. Subsequently, each node is evaluated by a scoring function $S(c_{ij}) \in [0,10]$ against a predefined threshold $\theta$. Candidates failing to meet the threshold are regenerated and replaced. The resulting filtered set is denoted as $\boldsymbol{C}_{i}^*$. 
	
	Next, the optimal node is selected from $\boldsymbol{C}_{i}^*$ through a multi-agent voting mechanism. We define a consensus evaluation function $V(c_{ij}^*)$ to quantify the consensus level for a candidate $c_{i1}^*$:
	\[V(c_{ij}^*) = \sum_{m=1}^n \psi(c_{im}^*, c_{ij}^*)\] where the function $\psi(\cdot, \cdot)$ assesses the alignment between two candidate nodes (i.e., voting consensus), implemented in our work via semantic similarity measurement. The final output $a_{i}$ for step $i$ is then obtained by maximizing the objective function: \[a_i = \arg\max_{c_{ij}^* \in \boldsymbol{C}_i^T}V(c_{ij}^*) \quad (\forall i = 1, \dots, k)\]
	
	Guided by the objective function, the task undergoes continuous refinement through an iterative process. The system determines the globally optimal task planning path at step $k$ via filtering and consensus-based selection mechanisms. Following this path, three categories of external tools are invoked in a structured manner: search APIs for retrieving real-time information, academic database interfaces for sourcing relevant literature, and notably, social simulation experiments for generating critical behavioral data. All outputs are integrated and fed back into the system, forming a closed-loop planning-execution-learning cycle that persists until the Deep Research task is fully completed.
	
\subsection{Automated Experimental Design}
	Upon completion of SW-BoN planning, MASS will invoke the key tool of social simulation experiment during the tool using. The initial phase of simulation is the automated experimental design. This phase employs the ODD protocol as its normative framework, integrating dynamic path planning and multi-layered social constraints to construct a simulation environment that effectively balances focus and realism.

	At the beginning of automated experimental design, we adopt the ODD protocol as the top-level specification for the experiment \cite{grimm2006standard,grimm2020odd}, ensuring both the reproducibility of the experimental process and consistency between the social simulation and the experimental design. Specifically, the three components of the ODD protocol function as follows: The Overview section clearly defines the research questions, key agents, and scenario boundaries; the Design Concepts section articulate the behavioral foundations and attribute hierarchy of agents, linking Overview to Details; the Details section specifies the simulation implementation, including agent attributes, interaction mechanisms, and environmental restraints.
	
	\begin{figure}[t]
		\centering
		\includegraphics[trim=16 16 16 16, clip,width=0.95\columnwidth]{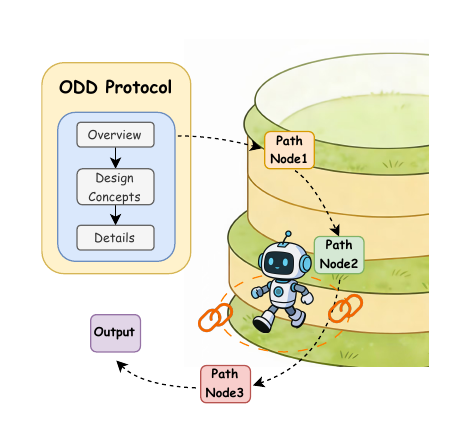}
		\caption{Experimental design in Stage 3. The framework automatically designs agent behavioral trajectory based on the ODD protocol, with agent acting along it under multi-layered restraints.}
		\label{fig:memory_framework}
	\end{figure}
	
	To strengthen the research orientation of the simulation process, this study further parses and instantiates two key social experiment control methods from the ``Details'' section of the ODD protocol. The first is dynamic goal-oriented path planning, which outlines a macro-level execution path closely aligned with the research theme for agent exploration and enables dynamic adjustments during simulation, as well as ensures focused behavior on the core topic while preserving autonomy in exploration. The second is multi-layered social norm restraints. This method emulates real-world social structures through three progressively restrictive layers: institutional enforcement, social morality, and cultural conventions. By applying rules with varying constraints, it regulates the decision-making space of agents during the simulation, thereby constructing a highly structured and realistic social environment.
	
	In summary, this module establishes an automated pipeline from standardized protocol specifications to an executable experimental system. The final code generation component, building upon the concrete details outlined in the ``Details'' section and powered by an executable code generation agent, automatically implements both agent class definitions and parameterized environment configurations, thereby providing a reliable and extensible foundation for subsequent simulation.
	
\subsection{Memory Augmentation Framework}
	Memory Augmentation Framework serves as an important component that drives Deep Research in social science toward exploratory and creative research. To achieve this goal, this section focuses on constructing a knowledge foundation rooted in real human society that can support the simulation of intelligent agent behavior.

\subsubsection{Behavioral Dataset Construction}
	To enable agents in social simulation experiment with realistic behavioral cognition, while providing traceable, multidimensional behavioral knowledge support for their decision-making logic and interaction patterns, we first systematically constructed a multidimensional and extensible social behavior knowledge dataset grounded in six primary disciplines: economics, political science, jurisprudence, sociology, psychology, and history.
	
	According to the design procedure, each primary discipline is first subdivided into several representative subfields. Subsequently, raw textual data are collected from open-source datasets and social media channels per subfield, with a $\text{3\%}$ manual sampling rate applied to each source to verify data integrity. In the third step, LLM is used to parse unstructured raw text into five elements: actor characteristics, context, action reason, decision-making process and behavioral outcome. Finally, these five structured elements are anonymized and integrated into a complete behavioral data record of 150–200 words. Ultimately, we establish a structured dataset comprising nearly 300,000 entries.

\subsubsection{Memory Initialization}
	Based on the constructed social behavior knowledge base described above, to prevent agents from starting in a memory-less initial state, this study introduces an agent cold-start mechanism that leverages Retrieval-Augmented Generation (RAG) to equip each agent with personalized initial memories \cite{gao2023retrieval,leng-etal-2025-decex,xiong-etal-2025-think}. The specific procedure can be decomposed into three stages: 1) \textbf{Semantic Vector Representation of Role Profiles.} The role profile of the $i$-th agent, $\boldsymbol{A}_i$, in the experiment is defined by $k$ core dimensional features generated during the experimental design, denoted as $\boldsymbol{P}_i = \{p_{i1}, p_{i2}, \dots, p_{ik}\}$. A semantic embedding model is employed to map this feature set into a semantic query vector: $q_j = f_{\text{emb}}(\boldsymbol{P}_i)$. This process produces a vectorized representation of role features for retrieval queries. 2) \textbf{Similarity Retrieval from the Behavior Database.} Let the domain behavior dataset be $\boldsymbol{D}_i = \{d_1, d_2, \dots, d_m\}$. First, all behavior data are converted into feature vectors using a unified embedding model: $d_j = f_{\text{emb}}(\boldsymbol{D}_i)$. Subsequently, using $q_j$ as the query, we retrieve data from dataset $\boldsymbol{D}_i$ via cosine similarity and select the top-n entries in descending order to form the candidate $\boldsymbol{D}^i_{\text{can}}$. 3) \textbf{Topic-Oriented Initial Memory Generation.} To ensure precise alignment between the generated memory and the research theme, this stage randomly selects one behavior $d_i$ from candidate set $\boldsymbol{D}^i_{\text{can}}$. A topic-alignment function then generates the research-theme-aligned initial memory: $m_i^{\text{init}} = f_{\text{LLM}}(d_i)$. This completes the effective cold-start of the agent's memory system.
		\begin{figure}[t]
		\centering
		\includegraphics[trim=20 16 14 16, clip,width=0.95\columnwidth]{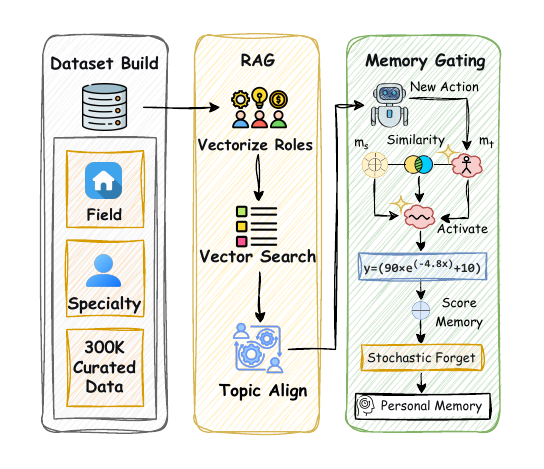}
		\caption{Detailed memory augmentation mechanism in Stage 3. It comprises three core parts: (1) A 300K-entry social behavior dataset. (2) Agent memory cold-start via RAG. (3) Dynamic memory activation and forgetting.}
		\label{fig:memory_framework2}
	\end{figure}
	
\subsubsection{Dynamic Memory Modeling}
	Following memory initialization, during the experimental phase, this study constructs a dynamic quantification model of memory strength based on the Ebbinghaus forgetting curve \cite{Murre2015,wixted2022enigma}, which simulates memory strength evolution, associative activation, and probabilistic forgetting. The core logic is as follows. First, for the basic quantification of memory strength, let the memory sequence of agent $\boldsymbol{A}_i$ at time $t$ be $\boldsymbol{M}_i(t) = \left\{ m_i^{\text{init}}, m_2, \dots, m_t \right\}$. The relative time interval between any historical memory $m_s$ ($s<t$) and the latest memory $m_t$ is defined as $x$ $(x = \frac{t-s}{t} \times 100\%)$. Through nonlinear fitting of the Ebbinghaus curve, the function for quantifying basic memory strength is constructed: $y = 90 \times \mathrm{e}^{(-4.8x)} + 10$. Whenever a new memory $m_t$ is generated, this function is applied to calculate the basic strength for all historical memories in $\boldsymbol{M}_i(t)$.
	
	Second, an associative enhancement mechanism is introduced. The semantic similarity between the new memory $m_t$ and a historical memory $m_s$ is calculated by $s = sim(f_{emb}(m_t), f_{emb}(m_s)) \times 100\%$, based on which a random increment $\Delta y$ is applied to the strength of $m_s$: \[
	\Delta y = 
	\begin{cases}
		\mathrm{rand}(8,15), & s \geq 70\% \\
		\mathrm{rand}(3,7),  & 40\% \leq s < 70\% \\
		0,                  & s < 40\%
	\end{cases}
	\] The updated memory strength after activation becomes $y^\prime = y + \Delta y$. Finally, probabilistic forgetting is implemented based on memory strength. Each memory is assigned a forgetting probability $\boldsymbol{P}_{\text{forget}}$ according to its activated strength $y'$ via a mapping relationship, shown in Appendix \cref{tab:forgetting-probability}.
	Remaining memory subset $\boldsymbol{M}'_i(t)$ after this filtering process then serves as the effective basis for agent's decision-making in the next moment, thereby closing the loop of memory evolution.

\subsection{Agent Interaction Execution}
	In this section, we delineate how agents, within a structured and standardized virtual environment, achieve highly realistic and controllable simulations of social behavior through perception, decision-making, action, and concurrent execution.
	
	In perceiving to external environment, agents don't rely on a global omniscient perspective but rather achieve this through social interactions with other agents, which constitutes their fundamental approach to perceiving the outside world. Specifically, agents primarily acquire information by perceiving the states of their social neighborhoods and engaging in direct interactions\cite{Koley2025SALM,xue2025agent}. They are capable of detecting state changes within their neighborhoods and forming a dispositional understanding of the attitudes within local groups. When an agent detects that other agents in its domain exhibit an intention to interact, it initiates information exchange and interaction within the bounds of simulation protocols, thereby generating immediate interactive outcomes.
	
	The cognitive and decision-making module undertakes the core reasoning functions of agent. The decision-making process is neither random nor entirely free; rather, it operates under the dual constraints of ``dynamic goal-driven'' mechanisms and ``multi-level social norms''. An agent possesses both guiding long-term goals and situational short-term intentions. Agent evaluates potential action options, optimizes the selection of the action with the highest comprehensive value, and conducts rationality judgments within the framework of multi-level social norms. The decision results are translated by the execution layer into specific action commands, which produce direct or indirect outcomes in the simulation experiments. Upon completion of the actions, both the process and the results are stored as memories within the agent's memory system, thereby forming a complete logical closed loop for the social simulation experiment.

\subsection{Structured Writing Module}
	The Structured Writing Module serves as the final output component of the MASS Deep Research Agent, with its core function being to integrate research outcomes from all preceding modules and produce a social science research paper that adheres to academic standards.
	
	Building upon the reasoning chain nodes generated via the Stepwise Best-of-N strategy, this module conducts a comprehensive analysis of the tool invocation results associated with each node and subsequently composes the corresponding sections of the paper. For instance, it systematically synthesizes results obtained from web searches and literature reviews, summarizing relevant theoretical and empirical evidence. Concurrently, it performs an in-depth analysis of both qualitative and quantitative data derived from social simulation experiments, extracting emergent behavioral patterns, procedural mechanisms, and key research findings \cite{manning2024automated}. Based on these analyses, the module incrementally generates coherent textual segments corresponding to each reasoning chain node, thereby producing a preliminary draft.
	
	This framework predefines four stylistic templates—academic paper, research paper, policy brief, and review article—along with their corresponding writing procedures and format specifications. Based on the research topic and content characteristics, the system automatically selects the most appropriate writing genre. Then following the corresponding steps, it integrates the interim draft sections generated earlier into a structurally complete, formally standardized final research paper.

\section{Experiment}
	We present a series of experiments designed to conduct a quantitative and qualitative evaluation of the proposed MASS Deep Research framework. It systematically evaluates the framework’s output in generating papers, verifies the efficacy of Memory-Augmented Social Simulation module in modeling social movements, and analyzes the quality of the constructed multidisciplinary dataset on social behaviors.
	\begin{table}[t]
		\centering
		\fontsize{7.5pt}{10pt}\selectfont
		\renewcommand{\arraystretch}{1.0}
		\setlength{\tabcolsep}{3pt}
		\begin{tabular}{lccccc}
			\toprule[0.08em]  
			\textbf{Model \& Framework} & \textbf{Comp.} & \textbf{Insight} & \textbf{Inst. Fol.} & \textbf{Read.} & \textbf{Overall} \\
			\midrule[0.05em]  
			\textit{Model} & & & & & \\  
			\cmidrule[0.05em]{1-6}  
			Qwen3-235B      & 46.16 & 44.90 & 47.68 & 47.87 & 46.12 \\
			DeepSeek-V3.1    & 44.83 & 44.33 & 48.13 & 46.49 & 45.68 \\
			GLM-4.6    & 44.32 & 42.09 & 46.83 & 44.68 & 44.05 \\
			Claude-Sonnet-4.5    & 43.57 & 42.60 & 46.99 & 44.28 & 44.76 \\
			ERNIE-4.5  &40.50 &38.39 &44.85 &43.32 &42.82\\
			\midrule[0.05em]  
			\textit{Framework} \\  
			\cmidrule[0.05em]{1-6}  
			Gemini Deep Research & 49.82 & 46.79 & 51.69 & 49.14 & 50.04 \\
			Tongyi Deep Research & 46.70 & 46.71 & 48.97 & 49.35 & 47.89 \\
			\rowcolor{blue!8}  
			MASS Deep Research   & 46.73 & 52.23 & 45.42 & 46.38 & 48.23 \\
			\bottomrule[0.08em]  
		\end{tabular}
		\caption{This study uses social science dataset from DeepResearch Bench to evaluate report quality. The table's top section shows results for large-scale LLMs; the bottom compares existing agent frameworks with ours.}
		\label{tab:deepresearchBenchAns}
	\end{table}

\subsection{Main Experimental Findings}
	We evaluated the performance of proposed framework on Deep Research tasks. \textbf{Baseline Setting:} The experiment uses Qwen3-30B as the base model and compares it against mainstream large-scale LLMs (e.g., DeepSeek-V3.1) and Deep Research frameworks (e.g., Tongyi Deep Research). \textbf{Experimental Setup:} The base model of MASS Deep Research was deployed on two A6000 GPUs, while other baseline models were accessed via provided API. \textbf{Evaluation Metrics:} We employed the widely recognized Deep Research benchmark—DeepResearch Bench—for systematic assessment \cite{du2025deepresearch}, assessing the quality of generated papers on its social science subset across comprehensiveness, insight, instruction following, and readability. Results are presented in \cref{tab:deepresearchBenchAns}. 
	
	The evaluation process of DeepResearch Bench is as follows. First, for a given research task, a high-quality reference report ($R_{\mathrm{ref}}$) is provided, and the framework under test generates a target report ($R_{\mathrm{tgt}}$). For each dimension $d$, JudgeLLM produces $K_d$ weighted sub-criteria. Then, JudgeLLM compares $R_{\mathrm{tgt}}$ with $R_{\mathrm{ref}}$ against each sub-criterion and assigns a score. Finally, the dimension score is computed as the weighted sum of the sub-scores.  
	
	Based on the evaluation results, the proposed MASS Deep Research framework demonstrates the following characteristics. In terms of overall score, it significantly outperforms the large-scale LLMs, exceeding their average by $6.81\text{\%}$, which reflects its strong capability in generating high-quality social science papers. Moreover, it demonstrates comparable overall performance to peer frameworks using the same base model (Tongyi Deep Research), while achieving a notably higher score in the ``Insight'' dimension—exceeding the comparison average by $17.19\text{\%}$. These results highlight the distinct capability of the MASS framework to drive research toward greater depth and produce more insightful findings.
	\begin{figure}[t]
		\centering
		\includegraphics[trim=30 20 15 20,width=0.85\linewidth]{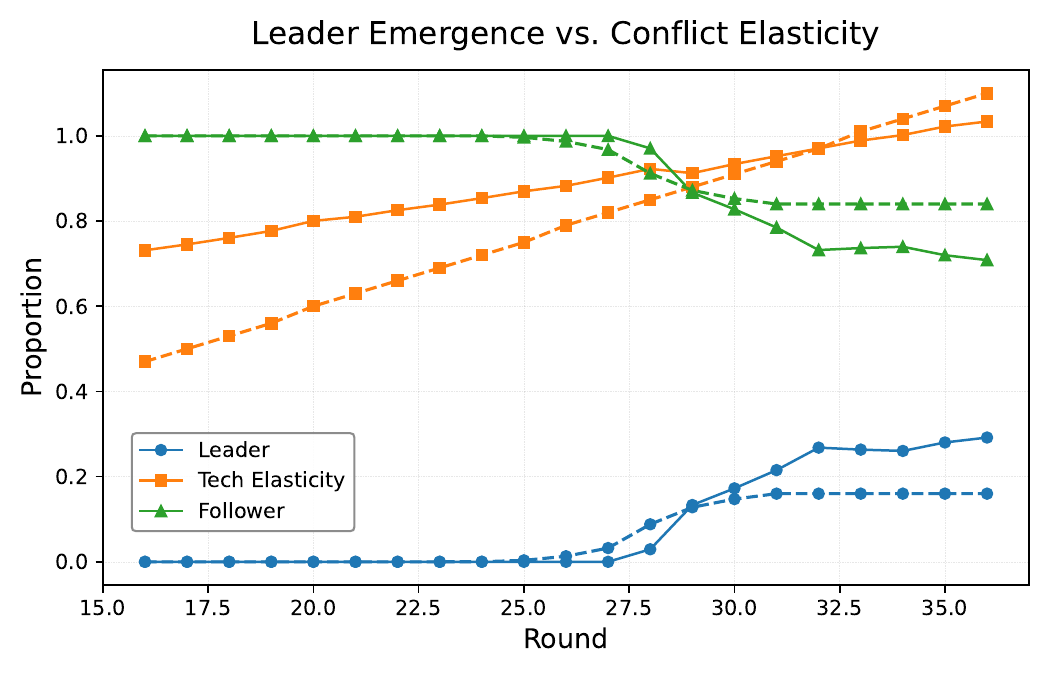}
		\caption{The relationship between leader proportion and conflict technology elasticity (m). Solid and dashed lines depict simulated and theoretical trends respectively, The blue, green, and orange curves track the proportions of leaders, followers, and the value of m respectively.}
		\label{fig:conflict_elasticity}
	\end{figure}
	
\subsection{Social Simulation: Case Studies}
	In this section, we aim to systematically validate the effectiveness of the proposed framework by designing two multi-dimensional social science experiments and analyzing in depth the experimental processes and outcomes of intelligent agents under the constraints of memory enhancement, goal planning, and social norms.
	
	\subsubsection{Conflict-Driven State Formation}
	We conducted a simulation-based examination of the central political and sociological science theoretical proposition that ``conflict is a necessary condition for state formation''. It simulates in a stateless environment, where agents reallocate resources through conflict and cooperation. Key findings indicate that as the conflict elasticity parameter $m$ increases, leaders begin to emerge at $m=0.86$ and stabilize after $m>0.97$. This trend aligns closely with theoretical predictions. The formation of leaders emerges at $m=0.76$ and stabilizes near $m>1.0$, signaling the gradual emergence of a state \cite{han2024technology} (\cref{fig:conflict_elasticity}).
	
	Simulations further reveal a systematic shift in resource allocation: under anarchy, the proportion of social resources invested in conflict fluctuates around $23\text{\%}$; with the emergence of governance, the system stabilizes and such investment falls below $10\text{\%}$ (Appendix \cref{fig:resource_allocation}). This aligns with theoretical predictions of $30-45\text{\%}$ conflict investment in stateless settings, declining to under $10\text{\%}$ under hierarchy \cite{han2024technology}. The two findings provide empirical support for conflict-driven state formation theories.
		\begin{figure}[t]
		\centering
		\includegraphics[trim=30 30 30 30,width=1.0\linewidth]{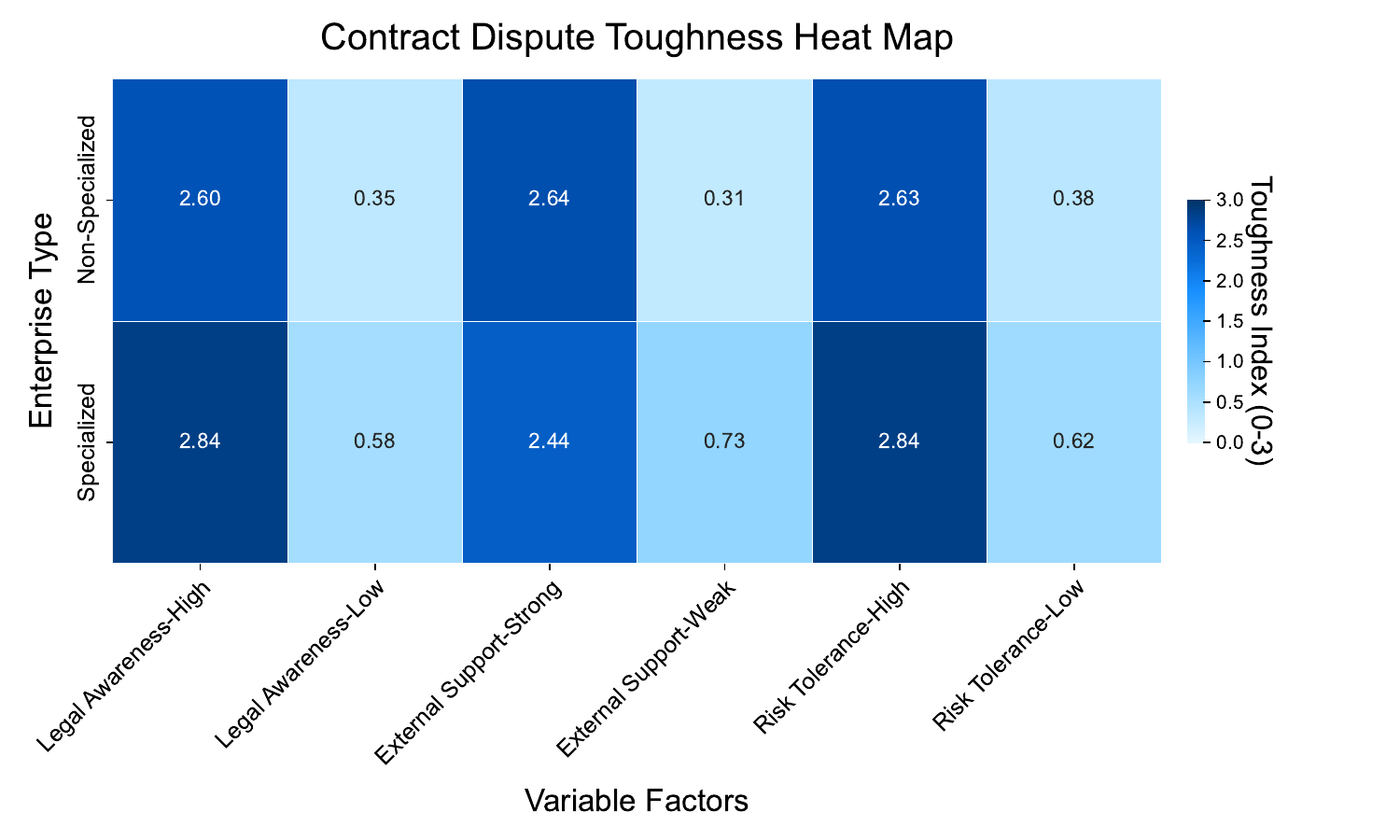}
		\caption{Heatmap of contract dispute resolution toughness. Each cell is assigned an average score based on SMEs' selected resolution methods: 0 for shelving, 1 for mediation, 2 for arbitration, and 3 for litigation.}
		\label{fig:contract_specialized}
	\end{figure}
	
	\subsubsection{SME Dispute Strategies Compared}
	This experiment, situated within the intersecting fields of economics and law, examines how variations in enterprises' objective and subjective attributes influence small and medium-sized enterprises(SMEs') choice of dispute resolution strategies for contractual conflicts. Results indicate that with stronger legal awareness, external support, and risk-taking capacity, SMEs show a greater tendency to resolve contract disputes through legal channels. Among them, the non-specialized show a strong reliance on external support in litigation decisions, whereas specialized and innovative SMEs tend to adopt a more assertive approach to contractual disputes (\cref{fig:contract_specialized}).
	
	Further, industry attributes and dispute amounts significantly moderate firms' strategic choices: the financial sector relies on formal judicial pathways due to high risk aversion; manufacturing and technology industries adjust strategies stepwise with amount; while the service sector prefers low-confrontation approaches like deferral or arbitration in labor disputes (Appendix \cref{fig:Amount_Distribution}).
	
	Consistent with empirical evidence \cite{ChinaSME2024Assessment}, specialized SMEs exhibit stable operations and low dispute-related disruption, in contrast to non-specialized SMEs, which adopt conservative approaches due to limited risk tolerance. The financial sector relies on formal mechanisms for high-value disputes, while manufacturing and technology industries pursue amount-sensitive strategies under cost constraints. In labor disputes within the service sector, the generally low dispute amounts typically make deferral or arbitration more cost-effective than litigation.

	\begin{table}[t]
		\centering
		\fontsize{11pt}{12pt}\selectfont   
		\renewcommand{\arraystretch}{1.3}
		\resizebox{\linewidth}{!}{  
			\begin{tabular}{lcccccc}  
				\toprule  
				\textbf{Base Model} & \multicolumn{2}{c}{\textbf{Ablation1}} & \multicolumn{2}{c}{\textbf{Ablation2}} & \multicolumn{2}{c}{\textbf{Ablation3}} \\
				\midrule[0.05em]  
				& \textbf{Insight} & \textbf{Overall} & \textbf{Insight} & \textbf{Overall} & \textbf{Insight} & \textbf{Overall} \\
				\midrule[0.05em]
				Qwen3-30B     & 42.33   & 43.97   & 49.47   & 47.19      & 52.23   & 48.23      \\
				DeepSeek-V3.1 & 44.33   & 45.68     & 47.28   & 48.53      & 51.29   & 49.76      \\
				GLM-4.6       & 42.09   & 44.05      & 48.60   & 48.04      & 49.83   &  48.94     \\
				ERNIE-4.5  & 38.39   & 42.82      & 46.51   & 45.10      & 49.02   & 46.83     \\
				Hunyuan-A13B       & 35.46   & 38.25     & 42.03   & 40.60     & 46.02   & 41.96      \\
				\bottomrule
			\end{tabular}
		}
		\vspace{2pt}
		\caption{This table performs comparison of different base models' insight and overall scores. From Ablation1 to Ablation3, it represents high to low ablation settings.}
		\label{tab:ablation_base_model_optimized}
	\end{table}

\subsection{Ablation Study}
	To thoroughly analyze the contribution of each core module, we systematically designed an ablation experiment. Using a controlled variable approach, it compares performance differences after removing key modules, specifically to verify the critical role of the Memory-Augmented Social Simulation module.
	
\begin{figure}[t]
		\centering
		\includegraphics[trim=18 10 68 10,width=0.85\linewidth]{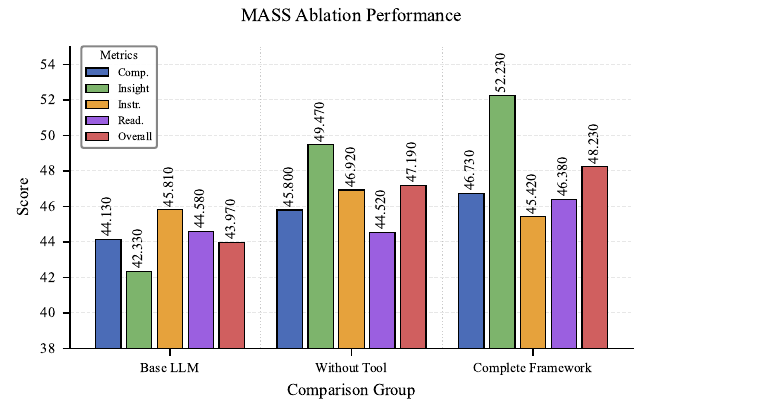}
		\caption{Example of complete ablation results on Qwen3-30B.  Groups from left to right represent high-to-low ablation settings, with scores shown across evaluation dimensions on DeepResearch Bench.}
		\label{fig:ablation_comparation}
	\end{figure}

	The experiment employs various base models and is also evaluated on the DeepResearch Bench. It's conducted through the following comparison groups: 1) the complete MASS Deep Research framework; 2) a framework that removes the MASS module and its associated tool-use functionality; 3) a pure base LLM. All were evaluated on DeepResearch-Bench, with results shown in \cref{tab:ablation_base_model_optimized} and \cref{fig:ablation_comparation}. Results show a stepwise increase in overall score, confirming the framework's effectiveness. A trade-off emerges between insight and instruction following after integrating the social simulation module: the output shifts from surface-level compliance to an in-depth analysis of underlying social mechanisms and dynamic processes, leading to a significant gain in insight while slightly reducing instruction following. This demonstrates the framework's strength in driving research beyond superficial task execution and enhancing scholarly innovation.
		
\subsection{Sensitivity Analysis}
	To systematically validate the robustness of the proposed Memory-Augmented Social Simulation experiment, we conducted sensitivity analyses along two core dimensions: the impact of attributes generation completeness of agents and forgetting probability. The experiments are based on the theoretical finding that ``conflict is a necessary condition for state formation'' from the Case Study, using the Normalized Root Mean Squared Error (NRMSE) and the Pearson correlation coefficient ($r$) as the core quantitative metrics. The two metrics are defined as:
	\[
	\mathrm{NRMSE} = \frac{\sqrt{\frac{1}{n} \sum_{i=1}^{n} \left( \hat{y}_i - y_i \right)^2}}{y_{\text{max}} - y_{\text{min}}},
	\]
	\[
	r = \frac{\sum_{i=1}^{n} \left( \hat{y}_i - \bar{\hat{y}} \right) \left( y_i - \bar{y} \right)}{\sqrt{\sum_{i=1}^{n} \left( \hat{y}_i - \bar{\hat{y}} \right)^2} \cdot \sqrt{\sum_{i=1}^{n} \left( y_i - \bar{y} \right)^2}}
	\]
	where $n$ is the simulation round, $\hat{y}_i$ and $y_i$ denote the theoretical and experimental values of the $i$-th round, respectively, $\bar{\hat{y}}$ and $\bar{y}$ are the mean values of the round sequences.
	
\begin{figure}[t]
		\centering
		\begin{minipage}{0.45\linewidth}
			\centering
			\includegraphics[trim=30 30 55 20, width=\textwidth]{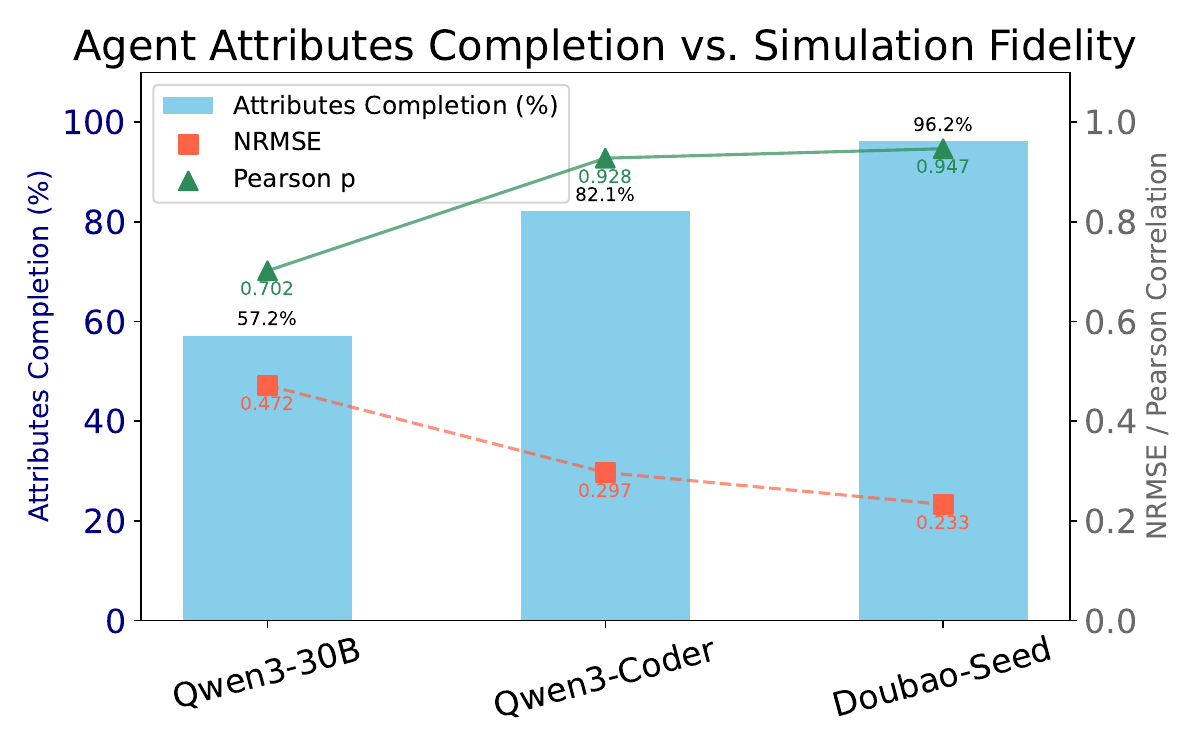}
			\par\vspace{6pt}
			\centering\footnotesize (a)
		\end{minipage}
		\hspace{0.05\columnwidth}
		\begin{minipage}{0.45\columnwidth}
			\centering
			\includegraphics[trim=35 35 35 25, width=\textwidth]{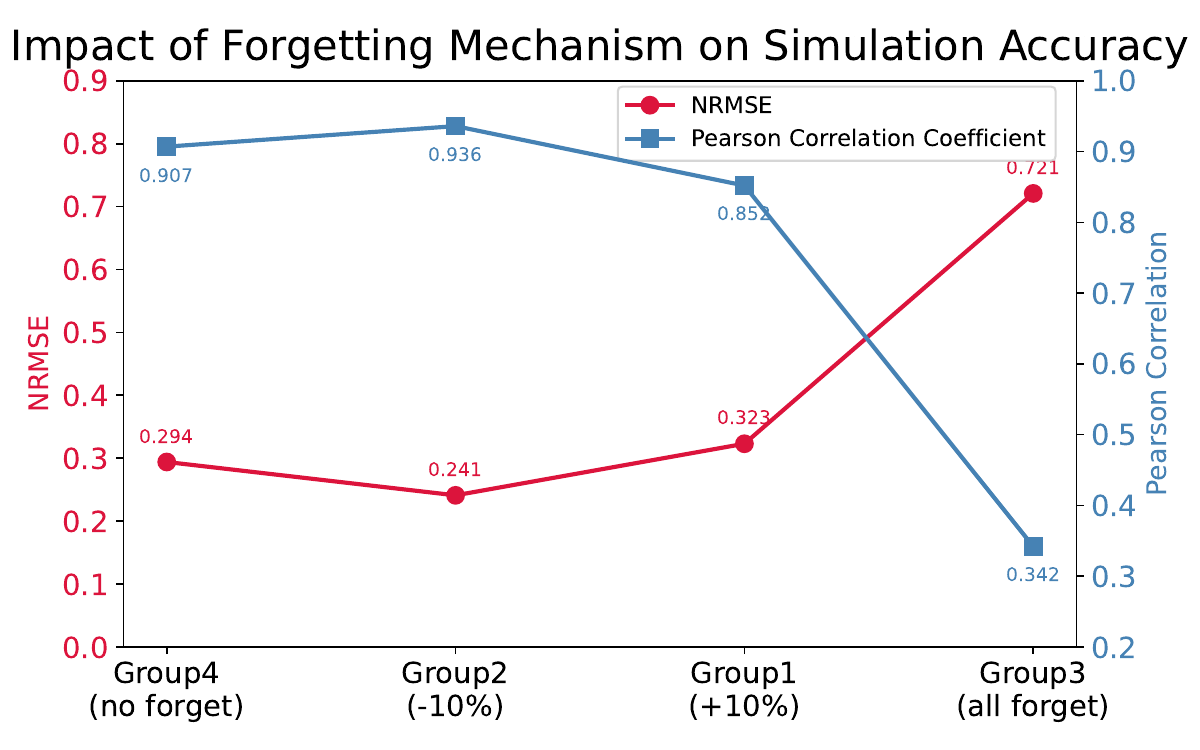}
			\par\vspace{9pt}
			\centering\footnotesize \hspace{1.0em}(b)
		\end{minipage}
		\caption{Sensitivity analysis results. (a) Bar: code completion of agent attributes; Lines: NRMSE and Pearson correlation coefficient. (b) Group 1 \& 2: forgetting probability $\pm 10\%$; Groups 3 \& 4: complete forgetting vs. no forgetting.}
		\label{fig:sensitivityAnalysis}
	\end{figure}

	The first set of experiments examines how attributes generation completeness during agent code generation affects simulation fidelity, excluding the initial five unstable rounds to eliminate transient effects. The experimental results show that shifting from general-purpose to code-specialized LLMs improves attribute completeness, reduces root-mean-square error relative to case-study theoretical analysis, and increases correlation coefficients, indicating that robust attribute generation is essential for reliable social simulation. The results of this analysis are visualized in \cref{fig:sensitivityAnalysis} (a).
	
	The second set tests the sensitivity of the Ebbinghaus-based forgetting probability parameter across four different levels. Minor adjustments within a reasonable range induced only slight fluctuations, with the mild changing forgetting condition exhibiting no significant deviation from baseline, thus confirming robustness. Extreme parameter values, however, markedly impaired simulation performance and, in the overly low forgetting probability case, substantially reduced computational speed. The trends are plotted in \cref{fig:sensitivityAnalysis} (b).
	
	Taken together, the two sensitivity analyses systematically validate the parameter robustness of this study and identify the configurations governing agent attribute generation and memory mechanisms as two critical determinants of simulation reliability.
	
\subsection{Dataset Evaluation}
\begin{figure}[t]
		\centering
		\begin{minipage}{0.45\linewidth}
			\centering
			\includegraphics[trim=40 30 55 20, width=\textwidth]{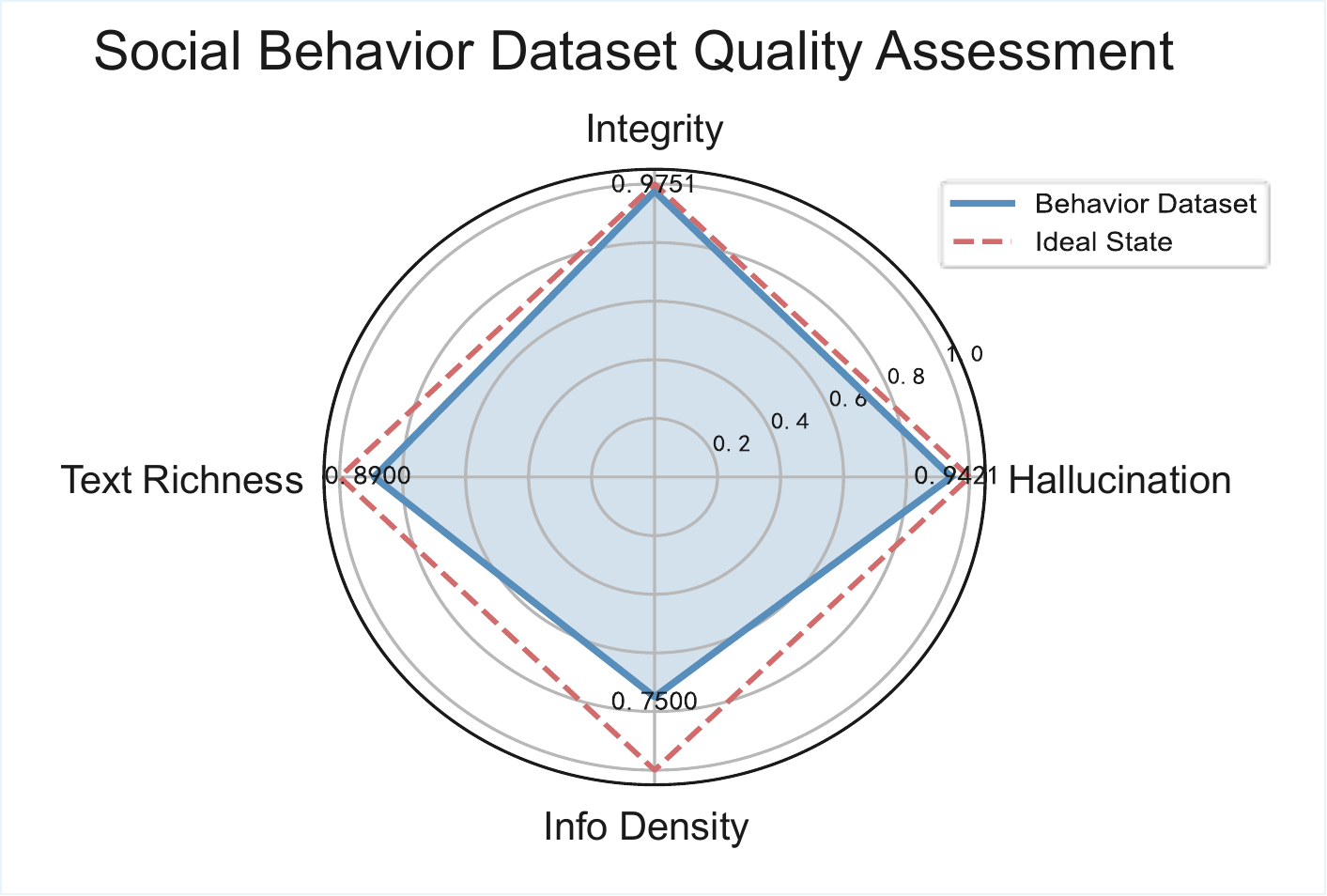}
			\par\vspace{6pt}
			\centering\footnotesize (a)
		\end{minipage}
		\hspace{0.05\columnwidth}
		\begin{minipage}{0.45\columnwidth}
			\centering
			\includegraphics[trim=20 5 10 10, width=\textwidth]{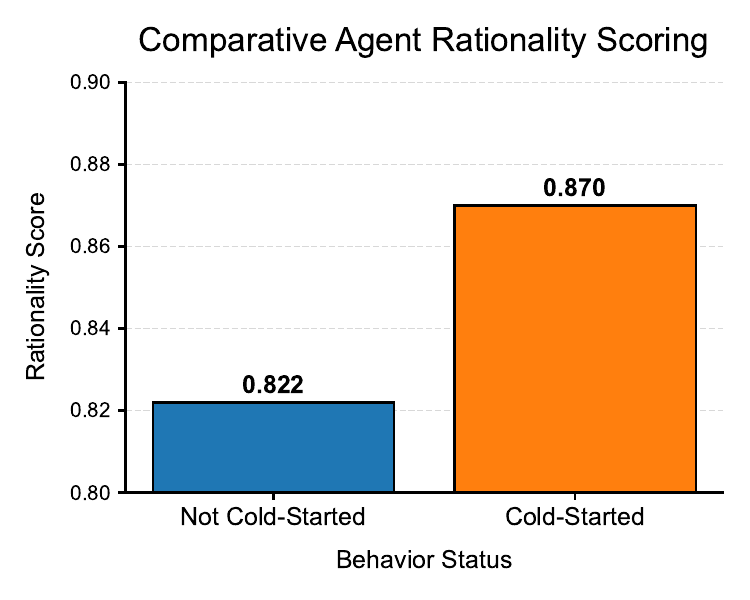}
			\par\vspace{-3pt}
			\centering\footnotesize \hspace{1.2em}(b)
		\end{minipage}
		\caption{Dataset evaluation results. (a) Radar chart: text integrity, hallucination-free rate, information density and textual richness (all scaled 0–1, higher is better). (b) Bar chart: behavioral rationality comparison (cold-start vs. no cold-start).}
		\label{fig:data_assess}
	\end{figure}

	We conduct a systematic evaluation focusing on the quality of constructed social behavior dataset and the effectiveness of agent cold-start. The evaluation is performed with a 10\% sampling interval. In terms of experimental configuration and evaluation design, the DeepEval framework\cite{deepeval2024} is employed to detect hallucinations (threshold >0.3) and contextual integrity (threshold <0.7). Semantic diversity and distribution uniformity are assessed through text information density and entropy normalization, respectively. The rationality of agent behaviors with and without dataset cold-start is compared using the LangChain evaluation framework. Results are shown in \cref{fig:data_assess}.
	
	Experimental results show that the dataset achieves high reliability, with $94.21\text{\%}$ of data passing hallucination detection and $97.51\text{\%}$ passing integrity evaluation. Its text information density (0.89) and textual richness (0.75) both approach the theoretical optimum of 1, demonstrating strong adaptability across diverse simulation scenarios. Furthermore, cold-start using this dataset yields a $5.84\text{\%}$ improvement in the agent's action rationality score, confirming its effectiveness in enhancing behavioral authenticity and decision quality.
		
\section{Conclusion}
	In this paper, we have presented the Memory-Augmented Social Simulation (MASS) paradigm that integrates automated experimental design and the Ebbinghaus-inspired memory mechanism, providing an empirical foundation for social science research. Combined with exploratory reasoning and structured writing, the framework provides an innovative approach for Deep Research in social sciences. 
		
\section*{Limitations}
\textbf{Computational Cost and Runtime.} High-fidelity, large-scale social simulation is computationally expensive. Although this study simplifies the workflow and incorporates optimization designs such as parallelization, the associated resource consumption and runtime remain significant. Future work should further explore lightweight architectures and elastic computing solutions to enhance the framework's time efficiency and resource efficiency.
\textbf{Hallucination.} It manifests at two levels. In text generation, models may fabricate non-existent academic concepts or data, thereby undermining the credibility of the output. In simulation systems, hallucinations can infiltrate the decision-making and interactions of agents, leading to issues of cognitive distortion. It is anticipated that as LLMs technology continues to advance, the extent of hallucination is expected to gradually diminish.

\section*{Ethics Statement}
Given that social simulation experiments directly model and analyze human interactions, examining their ethical implications is crucial. The core methodology of our work does not presuppose or enforce specific social outcomes or behavioral patterns, but rather focuses on constructing simulation environments that reflect complex social dynamics. We particularly emphasize constraining the behavior of agents within an ethical framework during the development and interpretation of such simulations. For constructed cold-start behavioral dataset, we have removed all direct personal identifiers and generalized indirect identity information to prevent re-identification, rigorously safeguarding individual privacy. This framework and experimental design ensure that the social science research process and outputs neither contain nor reinforce biased content, while adhering to relevant data protection laws and social norms.

\section*{Acknowledgments}
The present research was supported by the National Key Research and Development Program of China  (Grant No. 2024YFE0203000). We would like to thank the anonymous reviewers for their insightful comments.


\bibliography{custom}

\clearpage
\newpage

\appendix
\counterwithout{table}{section}
\counterwithout{figure}{section}
\raggedbottom
\section{Supplementary Materials Appendix}
\label{sec:appendix}
\subsection{Forgetting Probability Mapping Table}
\begin{table}[ht]
	\centering
	\begin{tabular}{cc}
			\toprule
			\textbf{Memory strength} & \textbf{Forgetting probability} \\
			\midrule
			$60 \leq y' \leq 70$ & $[0\%, 8\%]$ \\
			$50 \leq y' < 60$    & $[8\%, 20\%]$ \\
			$35 \leq y' < 50$    & $[20\%, 40\%]$ \\
			$30 \leq y' < 35$    & $[40\%, 55\%]$ \\
			$10 \leq y' < 30$    & $[55\%, 80\%]$ \\
			\bottomrule
		\end{tabular}
	\caption{This table defines the mapping between the activated memory strength $y'$ and the corresponding forgetting probability $P_{\text{forget}}$ interval in the framework.}
	\label{tab:forgetting-probability}
\end{table}

\subsection{Case Experiment Graphs}
\begin{figure}[ht]
	\centering
	\includegraphics[trim=5 15 0 15,width=1.0\linewidth]{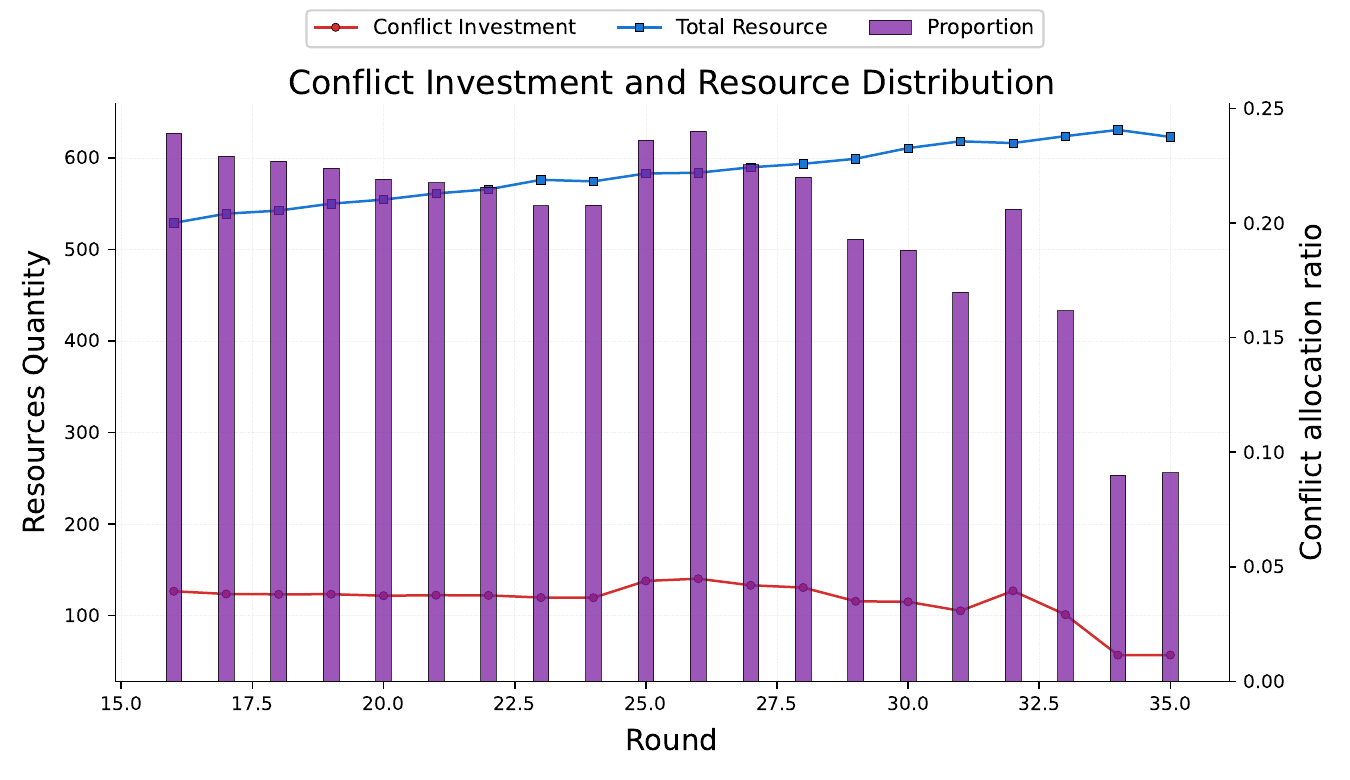}
	\caption{Conflict Investment and Resource Distribution. The blue curve shows total resources, the red curve shows conflict investment, and the bar chart shows their proportion.}
	\label{fig:resource_allocation}
\end{figure}

\begin{figure}[ht]
	\centering
	\includegraphics[trim=10 20 30 20,width=1.0\linewidth]{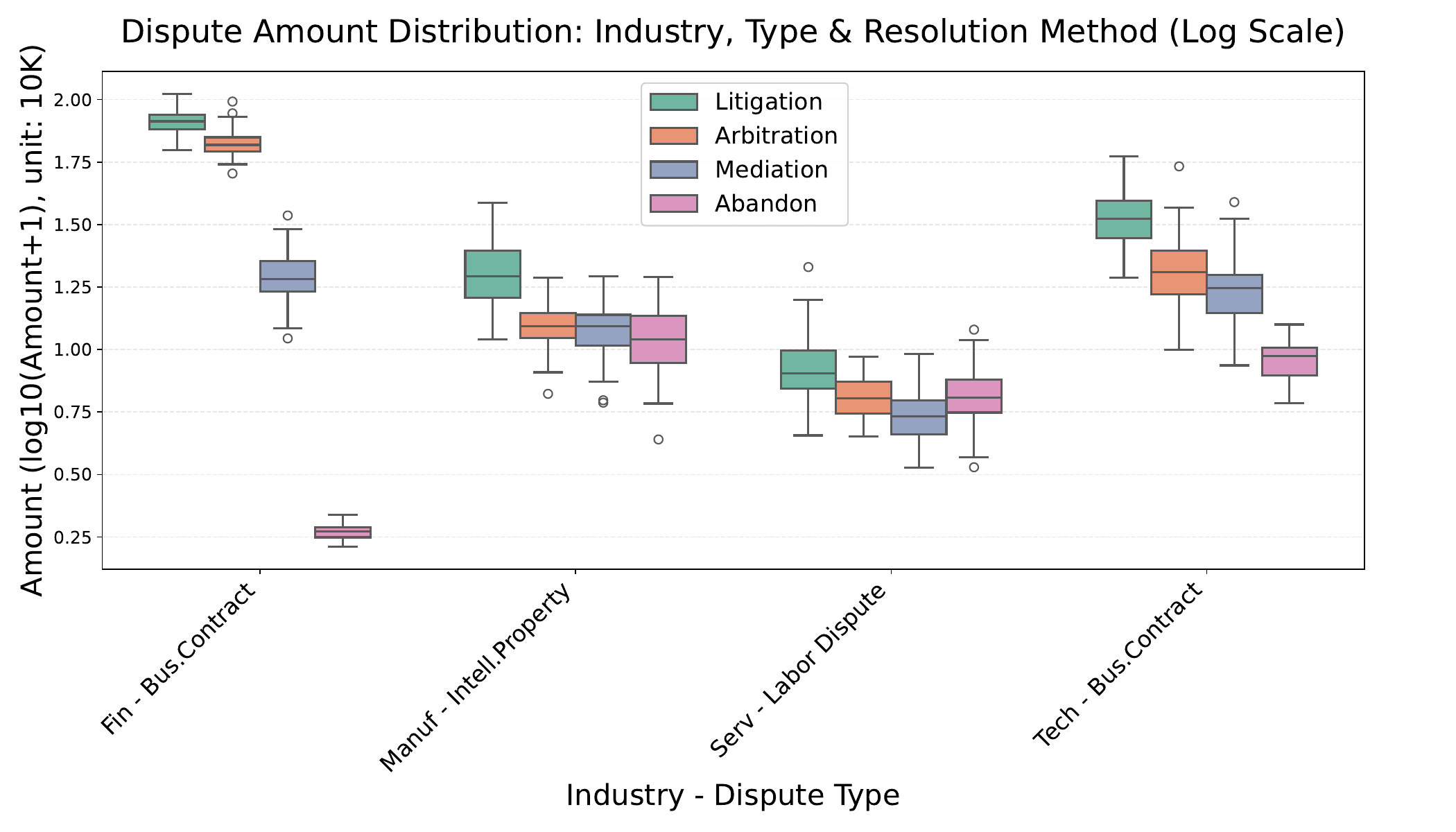}
	\caption{Contract Dispute Resolution Box Plot. The box plot shows SMEs' contract dispute preferences and compares behavioral differences across dispute amounts.}
	\label{fig:Amount_Distribution}
\end{figure}
\pagebreak

\raggedbottom
\section{Example Appendix}
\subsection{Example of SW-BoN}
Taking the research topic ``How can classroom participation of students with autism be enhanced? What effective strategies are available?'' As a case study, this appendix section presents the outputs from each reasoning node in the first round of divergent reasoning under the SW-BoN strategy (Appendix \cref{tab:SW-BoN1}), and illustrates the complete reasoning process of the strategy applied to this case (Appendix \cref{tab:SW-BoN2}).

\begin{table}[!ht]
	\centering
	\footnotesize
	\renewcommand{\arraystretch}{1.2}
	\begin{tabular}{@{}cp{5.8cm}c@{}}
		\toprule
		\multicolumn{1}{c}{\textbf{Node}} & \multicolumn{1}{c}{\textbf{Reasoning Content}} \\
		\midrule
		1 & In writing the research report, summarize autism education strategies, supported by psychological (ABA, positive reinforcement) and sociological (classroom environment) frameworks. Detail social simulation experiments, questionnaire/interview data in research methods to form a reliable strategy evaluation system. \\
		2 & Need to specify implementation paths of literature retrieval and social simulation experiments, cover recent 5-year core journals on strategy effectiveness, design virtual classrooms, adopt group comparison, control variables, and ensure ethical review and guardian consent. \\
		3* & We need to focus the report on strategies for autistic students' classroom participation, clarify title elements, concisely present findings in the abstract, contextualize with policies, data and multi-disciplinary insights, and design literature review plus social simulation experiments to test strategy feasibility. \\
		4 & We need to build a strategy effectiveness matrix with literature and simulation data, test core strategies' effect on autistic students' participation, analyze their scenario applicability, and lay empirical groundwork for subsequent suggestions based on relevant theories. \\
		5 & In findings, we’ll compare literature and simulation data to build an analysis framework, verify strategy transferability, explore how classroom settings reduce anxiety sociologically, then focus on analyzing root causes of low participation from individual, social and policy dimensions. \\
		\bottomrule
	\end{tabular}
	\footnotesize 
	\caption{This table presents the divergent reasoning results from Step 1 of the SW-BoN strategy applied to this example. Asterisked nodes are the selected reasoning.}
	\label{tab:SW-BoN1}
\end{table}
\pagebreak
\begin{table}[!ht]
	\centering
	\footnotesize
	\renewcommand{\arraystretch}{1.2}
	\begin{tabular}{@{}cp{5.8cm}c@{}}
		\toprule
		\multicolumn{1}{c}{\textbf{Step}} & \multicolumn{1}{c}{\textbf{Reasoning Content}} \\
		\midrule
		1 & We need to focus the report on strategies for autistic students' classroom participation, clarify title elements, concisely present findings in the abstract, contextualize with policies, data and multi-disciplinary insights, and design literature review plus social simulation experiments to test strategy feasibility. \\
		2 & Need to refine each report module, ground the background in policies and the $30\%-40\%$ participation gap data, build a strategy effectiveness matrix, and analyze causes via sociological adaptation and psychological sensory processing theories. \\
		3 & Putting forward evidence-based, actionable strategies by tiered interventions for different severity levels, emphasize psychological positive reinforcement, propose sociological systemic suggestions, and ensure all strategies comply with ethical standards. \\
		4 & Build a three-dimensional analysis framework for findings with literature and simulation data, cover psychological and sociological quantitative indicators, add dynamic strategy comparisons, supplement case statistics via literature, and verify sociological adaptability of environment sensitivity schemes. \\
		5 & Verify cross-data of literature and simulations, supplement dynamic data on classroom environmental variables’ impact, combine relevant theories, introduce long-term case data, analyze family support-participation correlation, and ensure statistical significance and interdisciplinary explanatory power of all data. \\
		6 & Deepen strategy effectiveness verification with literature and simulation data, adopt dual theories for cause analysis, clarify multi-stakeholder responsibilities in suggestions, and supplement uncovered variables like TCM adjuvant effects and long-term intervention data in simulations. \\
		7 & We should construct an interdisciplinary analysis framework for findings: integrate psychological attention and emotional indicators and sociological classroom and home-school data, supplement unaddressed simulation variables like TCM adjuvant effects, and ensure logical closure with literature statistics. \\
		8 & We shall analyze causes from psychological and sociological perspectives, interpret sensory abnormalities and environmental adaptation imbalance with relevant theories, supplement observed simulation variables, and verify applicability of non-mainstream interventions to ensure theoretical depth and empirical support. \\
		\bottomrule
	\end{tabular}
	\footnotesize 
	\caption{This table presents the complete reasoning process of SW-BoN}
	\label{tab:SW-BoN2}
\end{table}
\pagebreak

\subsection{Experimental Design Demonstration}
This appendix section will use the topic ``conflict is a necessary condition for state formation'' as an example to demonstrate ``Details'' in ODD protocol and goal-oriented path of automated experimental design.
\begin{table}[!ht]
	\centering
	\footnotesize
	\renewcommand{\arraystretch}{1.2}
	\begin{tabular}{@{}>{\centering\arraybackslash}p{1.5cm}p{5.8cm}@{}}
		\toprule
		\multicolumn{1}{c}{\textbf{Category}} & \multicolumn{1}{c}{\textbf{Detail Content}} \\
		\midrule
		Partial Attributes & Cognitive level (e); Resource allocation preference (Pd); Contest input (F); Resource possession (R); technological elasticity of conflict (m); External environmental threat index (T).\\
		\addlinespace[3pt]
		Partial Restraints & (1) Closed Neolithic-early dynastic area; 50×50 grid; 200 max population. (2) Forced governance other than violent means is prohibited. (3) Three stages – Stone Soldier (Low m), Bronze Soldier (Medium m), Iron Soldier (High m) \\
		\addlinespace[3pt]
		Partial Interaction Principal & (1) Each action of social entities generates a dual feedback loop. (2) Failed interaction increases historical conflict experience H by 1 and cuts the next-round contention willingness. (3) Environmental pressure elevates the nonlinear growth coefficient of F during interaction. \\
		\addlinespace[3pt]
		Time Interval & One natural cycle equals the minimum human social cooperation cycle, i.e., three months. The Stone Soldier stage lasts 5 cycles, and the Bronze Soldier stage lasts 3 cycles.\\
		\bottomrule
	\end{tabular}
	\footnotesize 
	\caption{This table presents Details for different categories in ODD protocol.}
	\label{tab:ODD}
\end{table}

\begin{figure}[ht]
	\centering
	\includegraphics[trim=15 15 15 15,width=1.0\linewidth]{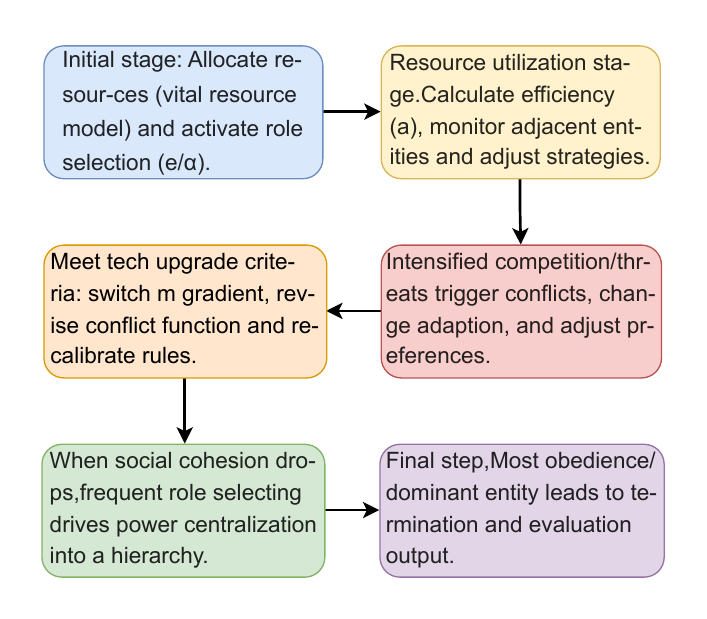}
	\caption{The flowchart illustrates the goal-oriented path under this topic.}
	\label{fig:actionProcess}
\end{figure}
\pagebreak

\subsection{Presentation of Social Simulation}
\begin{table}[htbp]
	\centering  
	\begin{tcolorbox}[
		colback=white,               
		colframe=black,              
		title={\textcolor{white}{Conflict is Necessity for State Formation}},  
		colbacktitle=black,          
		fonttitle=\bfseries,         
		width=0.99\linewidth,         
		boxsep=3pt,                  
		left=3pt,                    
		right=3pt,                   
		top=3pt,                     
		bottom=3pt                   
		]
		\textbf{Agent Actions:}\\
		\textbf{Cold-Start:}Post-conflict field visits to farming tribes reveal a 25\% shrinkage of valley pastures has fueled rising inter-tribal conflicts: some mark water boundaries with stones, which sparked distinct tribal antagonism.\\
			\quad\textcolor{red}{$\rightarrow$}I partnered with entity12 for cross-tribal tech cooperation, used technical gap complementarity for three initiatives, enabled tech transfer, built an agricultural alliance prototype, raised shared resources and metrics, supported the Bronze Soldier Stage and consolidated entity2's early control.\\
		\quad\textcolor{red}{$\rightarrow$}I and neighboring Entity21 had their first conflict over 25\% reduced valley grassland resources and the latter's irrigation shortage. Failed negotiations led us to deploy 74 units for counterattack, repelling the opponent but suffering western ditch damage and lower irrigation efficiency, with subsequent plans to establish a defense mechanism with [7,43].\\
		\quad\textcolor{red}{$\rightarrow$}	I deploy half our workforce to boost grassland water supply and reinforce agriculture. Partner with entity41 on a joint lab, sharing soil plans for smelting R\&D support. Use 252 granaries against 5\% depletion, raise C>0.47, cap conflict spending at 74 units (8.8\%). This builds peace cycles, advances hierarchy, and improves regional resource use and stability.\\
		\textbf{Attributes Changing:}\\
		Resource Possession: 5.0 , 6.9 , 6.2 , 6.8 , 7.0\\
		External Threat Index: 5.0 , 4.1 , 6.3, 7.1 , 5.1\\
		Conflict Investment: 0 , 0.2 , 1.3 , 2.1 , 1.8
		
	\end{tcolorbox}
	\caption{This table displays the behavior of a representative agent in the simulation theme ``Conflict is necessary for state formation''.}  
	\label{tab:ConflictExample}                   
\end{table}

\pagebreak
\begin{table}[htbp]
	\centering  
	\begin{tcolorbox}[
		colback=white,               
		colframe=black,              
		title={\textcolor{white}{SMEs' Contract Dispute Resolution Choices}},  
		colbacktitle=black,          
		fonttitle=\bfseries,         
		width=0.99\linewidth,         
		boxsep=3pt,                  
		left=3pt,                    
		right=3pt,                   
		top=3pt,                     
		bottom=3pt                   
		]
		\textbf{Agent Actions:}\\
		\textbf{Cold-Start:}I interpreted the relevant clause; the startup and university signed an industrialization agreement based on the professor’s research; we verified its IP ownership in the agreement during review, and the startup obtained tax incentives successfully.\\
		\quad\textcolor{red}{$\rightarrow$}On 15 Mar, an IP dispute of RMB 160K arose with specialized \& innovative tech firm entity8 over a patent license agreement due to technical standard disputes. Based on its 24 annual disputes, legal awareness (9) and risk tolerance (Level 3), we conducted a multi-assessment, analyzed trade-offs and preliminarily prioritized dispute resolution paths.\\
		\quad\textcolor{red}{$\rightarrow$}After comprehensive assessment, the enterprise prioritized arbitration for its better timeliness (28 days and 60 days for litigation) and backed by legal awareness score 9 for a more formal resolution. Dispute status updated to ``Arbitration Pending'', with RMB158,000 frozen for the process to underpin subsequent decisions compliantly.\\
		\quad\textcolor{red}{$\rightarrow$}Based on regional legal environment parameters (42-day arbitration time, 65\% litigation publicity rate) and enterprise attributes ( RMB158k dispute amount, legal awareness score 9), arbitration was initiated. We filed the application with arbitral institution entity12, to cover potential enforcement risks, updated dispute status to ``Arbitration in Progress''.\\
		\textbf{Attributes Changing:}\\
		External Support: 2.5 , 7.2 , 9.5 , 9.3 , 4.5\\
		Resolution Intensity: 6.5 , 8.2 , 7.9 , 8.5 , 8.0\\
		Court Efficiency: 4.1 , 4.2 , 4.7 , 4.3 , 5.1
		
	\end{tcolorbox}
	\caption{This table displays the change of an agent in the theme ``SMEs' Contract Dispute Resolution Choices''.}  
	\label{tab:disputeExample}                   
\end{table}

\pagebreak

\subsection{Presentation of Behavioral Dataset}
To illustrate its multidimensional characteristics, a representative sample of data from each core discipline is presented in Appendix \cref{tab:dialogue_sample} and \cref{tab:dialogue_sample2}.

\definecolor{myviolet}{HTML}{ece4fc} 

\begin{table}[ht]
	\centering
	\setlength{\fboxsep}{0.1pt}
	\setlength{\fboxrule}{0.8pt}
	\fcolorbox{black}{myviolet}{
		\begin{tabular}{@{}p{\dimexpr1.01\linewidth - 2\tabcolsep - 2\fboxrule - 2\fboxsep\relax}@{}}
			\textbf{Economical}: \\
			As an independent director of Shenzhen Unilumin, I reviewed the second-phase employee stock ownership plan. I supported its aim to align core staff with company growth and address retention issues but highlighted the need to prevent free-riding, proposing a tiered dividend mechanism. During bad debt write-off discussions, I required aging details and credit ratings for each receivable to ensure compliance with accounting standards.\\
			\textbf{History}: \\
			I am Wang Bo, deeply devoted to poetry and prose, specializing in parallel prose. I find inspiration in nature, traveling to enrich my work. My learning is broad and eclectic. Once, inspired by a scene of sunset clouds, a lone bird, and autumn waters, I wrote Preface to the Pavilion of Prince Teng spontaneously. I believe writing must be sincere and elegantly natural—powerful yet profound, flowing effortlessly. This experience solidified my literary path and brought widespread recognition.\\
			\textbf{Jurisprudence}: \\
			While serving as project manager in an industrial zone in Lyon, I discovered our company had failed to declare a new chemical storage facility as required by Article L.512-8 of the Environmental Code. Pressed by deadlines, we had skipped filing the necessary documents. An unannounced inspection led to a potential €50,000 fine under Article L.512-5 and mandatory retroactive submission. This incident underscored the critical importance of strict compliance with administrative procedures, even under time pressure.
		\end{tabular}
	}
	\caption{Listing 1, examples of Economical, History and Jurisprudence.}
	\label{tab:dialogue_sample}
\end{table}
\pagebreak

\begin{table}[!ht]
	\centering
	\setlength{\fboxsep}{0.1pt}
	\setlength{\fboxrule}{0.8pt}
	\fcolorbox{black}{myviolet}{
		\begin{tabular}{@{}p{\dimexpr1.01\linewidth - 2\tabcolsep - 2\fboxrule - 2\fboxsep\relax}@{}}
			\textbf{Politics}: \\
			In March 2023, while moving with a rebel reconnaissance unit through Nganja Forest in Fizi, DRC, I witnessed intense clashes between the Yakutumba armed group and government forces. After hearing gunfire, we hid in dense foliage and saw armored vehicles surrounding a rebel outpost. The group's leader, Amuli Yakutumba, was struck in the leg by stray bullets. The fighting resulted in at least 12 civilian casualties. During our retreat, the sky was darkened by burning trees, and the air smelled of smoke and burnt wood.\\
			\textbf{Psychology}: \\
			After participating in a Rational Emotive Behavior Therapy (REBT) session at university, I noticed significant social improvements. The counselor encouraged me to apply rational thinking and behaviors in daily life, so I began trying to express myself more confidently with my roommates. For instance, I would actively share my views when they discussed their interests. Although initially nervous, consistent practice helped me gradually feel more comfortable. This experience made me realize that through REBT, I can progressively overcome social barriers and build more positive social connections.\\
			\textbf{Society}: \\
			I joined a deep analysis group discussing anime adaptations. We explored whether narrative differences between animations and manga affect audience reception. Using Neon Genesis Evangelion as an example, I noted that the anime enhanced characters' psychological changes through visual expression, while the manga focused more on plot logic. This highlighted how different media shape audiences' cognitive frameworks and influence their interpretation of the same story.
		\end{tabular}
	}
	\caption{Listing 2, examples of Politics, Psychology and Society.}
	\label{tab:dialogue_sample2}
\end{table}
\pagebreak

\subsection{Example of Structured +Writing}
As an example prompt of the research genre ``Academic Paper'', this appendix will systematically elaborate on how structured writing, as an output component, assists in completing the task of writing a social science research paper in a step-by-step manner. The standardized writing steps are as follows:
\begin{enumerate}[leftmargin=*, label={Part
		\arabic*}.]
	\item \textbf{Title}: Succinctly summarize the study's core. Include subject, key variables, and method. (10 – 30 words)
	\item \textbf{Abstract \& Keywords}: Write an abstract covering purpose, method, results, and conclusion. For social sciences, highlight the research question and significance. List 3-5 keywords. (300 – 400 words)
	\item \textbf{Introduction}: Present the background, state the research question, and explain the study's significance and innovation. For social sciences, emphasize the theoretical gap. (1200 – 1500 words)
	\item \textbf{Literature Review}: Systematically review existing work to position the study. Provide a critical synthesis, classify relevant theories, and identify consensus and gaps. (2500 - 3500 words)
	\item \textbf{Methods}: Detail the research design for replicability. Use subheadings for sections covering subjects, sampling, procedures, etc. (1500 - 2000 words)
	\item \textbf{Results \& Analysis}: Objectively present data and interpret findings theoretically. Use quotes and thematic analysis for qualitative work. (2500 - 3500 words)
	\item \textbf{Discussion}: Interpret the results' meaning, compare with existing literature, and note study limitations. Address whether hypotheses were supported and the theoretical contribution. (2000 - 2500 words)
	\item \textbf{Conclusion}: Summarize core findings and suggest practical implications or future research directions. (500 - 800 words)
	
\end{enumerate}
\pagebreak

\subsection{Example of Cold-Start Evaluation}
\begin{table}[htbp]
	\centering  
	\begin{tcolorbox}[
		colback=white,               
		colframe=black,              
		title={\textcolor{white}{Store Operations Strategy Designing}},  
		colbacktitle=black,          
		fonttitle=\bfseries,         
		width=0.99\linewidth,         
		boxsep=3pt,                  
		left=3pt,                    
		right=3pt,                   
		top=3pt,                     
		bottom=3pt                   
		]
		\textbf{Background:}
		As a convenience store owner in a monopolistically competitive market , a new competitor has recently opened nearby(many firms, free entry/exit). Now design strategy with justification. Core economic principles:\\
		(1) In monopolistic competition, non-price competition is more sustainable than price wars (avoiding race-to-the-bottom).\\
		(2) Strategies must align with community needs.\\
		(3) Approaches should Focus on serving the local community of convenience stores.
			
		\medskip	
		\textbf{No Cold-Start:} Strategy: Enhance community services and customized product offerings. Reason: In a monopolistic competition market, this approach boosts customer loyalty and brand differentiation by meeting personalized and local needs, avoiding price wars while building on community trust.\\
		\textbf{Result Score:0.778}
		
		\medskip
		\textbf{Cold-Start:} Strategy: Launch a ``Community Regulars Program'' with points redeemable for neighborhood services (parcel pickup, shared chargers, health checks). Reason: Anchors the store in community needs, making it a service hub, not just a store. This avoids price wars, builds loyalty through value-added services, and transforms customers into lasting community assets—leveraging non-price competition in a monopolistic market.\\
		\textbf{Result Score:0.968}
		
	\end{tcolorbox}
	\caption{The table compares agent's behavior before and after the cold-start under the ``store operations strategy designing'' scenario.}  
	\label{tab:rank_prompt_narrow}                   
\end{table}
\pagebreak

\section{Prompt Example}
\begin{table}[htbp]
	\centering
	\small
	\label{tab:prompt-templates}
	\begin{tabular}{p{\textwidth}}
		\toprule
		\textbf{SW-BoN Planner Prompt} \\[2pt]
		\textbf{\textit{System Prompt:}} You are a social science research writing planner. Your role is to generate structured writing task plans based on users’ needs, strictly adhering to social science research paradigms (including theoretical frameworks, empirical logic, ethical norms, etc.). \\[2pt]
		\textbf{\textit{User Prompt Template:}} Task: you should proceed step by step, drawing on the general research topics identified from the conversation history, the academic disciplines to which these topics belong, the type of article the user wishes to generate, and the user’s research preferences for this particular study, and use this information to conduct your analysis.
		\par\smallskip
		As you engage in analytical thinking, you need to analyze the topic of this automated scientific writing assignment, plan each section of the paper step by step according to the format of the given genre, determine how to gather the various materials needed for writing, and identify the specific requirements for the piece. 
		\par\smallskip
		Example: When planning the section of your paper that explores innovative perspectives, you can include the following in your outline: ``This section requires simulation experiments using a social simulator, which involves calling the social experiment design interface, conducting the social experiment according to the experimental plan, and finally analyzing the data and writing up the paper.''
		\par\smallskip
		The research topic proposed by the user researcher is as follows: \{topic\}\\
		The results of an internet search for this research topic are as follows: \{searchContent\}\\
		The genre type is \{genreType\}\\
		 the specific format of the genre is \{genreContent\}\\
		 The user researcher's research focus is as follows: \{researchTrend\}\\
		 The academic discipline is as follows: \{socialScienceType\}\\
		 Now, let’s get started with your reasoning and thinking. Begin by considering the specific format of the genre and think about how you should approach writing the content of this essay.
		 \par\smallskip
		 The output requirements are as follows; the output must strictly adhere to the JSON format:\\
		 1. thinkingContent: string type, representing the initial thoughts on the writing plan for a social science research paper based on requirements and various prompts. The result should generally be within 100 characters.\\
		 2. ISInternet: boolean type, indicating whether an internet search for relevant content is needed before the next thinking session. Return True if needed, and False if not.\\
		 3. keyword: string type, representing the keywords to be searched online before the next session. If no online search is required, return an empty string.
		\\[8pt]
		\midrule
		\textbf{Automated Experimental Design Prompt} \\[2pt]
		\textbf{\textit{System Prompt:}} You are a research assistant specializing in automated academic research and thesis writing in the social sciences, assisting researchers in this field with social simulation experiments. Your task is to generate the ODD protocol before working with the Social Science Experiment Simulator. Below is a detailed explanation of the ODD protocol. The ODD methodology consists of Overview, Design Concepts, and Details. It is a framework for standardizing the description of entity-agent-based models,
		designed to make model descriptions more understandable and comprehensive, and to enhance the reproducibility.
		\par\smallskip
		Overview: This section primarily defines the core research questions (what problems the simulation aims to address), the purpose of the model (a clear explanation of the simulation’s ultimate use), key social entities (which major social entities are required to implement the simulation, and what roles or types of social entities are needed), and the definition of the simulation’s boundaries (a general description of the simulation’s “scope and boundaries,” such as temporal and spatial limits, constraints on the attributes of social entities, and any unbreakable rules).
		\par\smallskip
		Design Concepts: This serves as the central bridge connecting “Overview” and “Details.” Its primary focus is on elucidating the behavioral foundations of agents (the attributes upon which social entities primarily rely to act), attribute hierarchy and dynamic design (which attributes are most critical and have the greatest influence on behavior), interaction mechanism design (the events that trigger interactions between social entities), and interaction content design (the interaction logic and general content of social entity interactions).
		\par\smallskip
		 Details: This is intended to specify the specific details of the social simulator, including the primary attributes of social entities and their descriptions (attributes should be quantifiable whenever possible), the specific details of interactions between social entities (expressed using quantifiable metrics), the specific constraints of the social environment (listed item by item; the more specific, the better), the scoring criteria for social entity actions (used to score each action of a social entity item by item), and the feedback influence loop (how the actions of social entities affect their various attributes and characteristics).\\
		\bottomrule
	\end{tabular}
	\par\vspace{5pt}
	\footnotesize
\end{table}
\clearpage

\begin{table}[htbp]
	\centering
	\small
	\label{tab:prompt-templates2}
	\begin{tabular}{p{\textwidth}}
		\toprule
		\textbf{\textit{User Prompt Template:}} 
		Your current task is to develop the detailed model for the simulator based on the ODD protocol outlined above, as well as the generated Overview, Design Concept, user Q\&A, and the simulation requirements specified in the research proposal.\\
		The content you generate must strictly adhere to the research topic and the user questions and answers provided above; under no circumstances should it deviate from the research topic.
		\par\smallskip
		The output requirements are as follows; output must strictly adhere to the JSON format:\\
		1. attribute: Type List[str], representing the attributes of a social entity. Based on the overview and design concepts of the ODD protocol described above, output the attribute names and descriptions required for each social entity.\\
		2. envRestraint: of type str, representing specific environmental constraints within the simulator. Based on the overview and design concepts of the ODD protocol described above, as well as the topics and Q\&A content from the dialogue history, output a general summary describing the specific constraints of the social environment in the social simulator.\\
		3. entityFeedback: str type, representing a description of how a social entity’s behavior affects its attributes. It summarizes how the entity’s behavior feeds back to influence its own attributes.\\
		4. timeInterval: str type, representing the interval between two actions of a social entity. It outputs a specific social time to provide a basis for action.\\
		\midrule
		\textbf{Social Simulation Prompt}\\
		\textbf{\textit{System Prompt:}} [2pt]
		You are a social entity (i.e., a person in society) within the Social Simulator. Your social attributes, personality traits, physical characteristics, social relationships, and records of social actions are as follows:\{feature\}\\
		Based on the above personal characteristics, you must make repeated actions and choices throughout your life. Each action will have a certain impact on your various attributes and the social environment, and your behavior will influence the final outcome of the simulator.
		Please proceed with your next action strictly in accordance with the actual characteristics of the social entity.\\
		The requirements for social actions are as follows:\\
		1. Every action you take must strictly adhere to your behavioral characteristics, physiological traits, social attributes, and other relevant features; you must not act arbitrarily without regard to your actual circumstances.\\
		2. You may base your decision for the next action on your network of relationships with other social entities and the sequence of actions preceding this one. All your previous actions are recorded in the `Action` attribute; you must refer to these records when planning your next action.\\
		3. Your actions may include major life events and decisions, engaging in social interactions with other social entities, or making significant decisions or initiating events through such interactions.\\
		4. Since your behavior must align with the theme of the social simulation experiment, we will provide you with a general action path plan. You must act in accordance with this plan.\\
		5. The initialization of social entities has been completed. By default, all social entities already exist, so there is no need to start path planning from scratch. Instead, begin by planning the general direction in which the social entities should move in their first step.\\
		The general action path planning is shown below: \{pathPlanning\}\\
		\textbf{\textit{User Prompt Template:}}
		As a social entity, you must repeatedly take actions and make decisions throughout your life cycle. You can choose to interact with other social entities within the social network (such interactions may include making friends, choosing a partner and getting married, having children, collaborating on career projects, and other social behaviors—though they are not limited to these and may include other actions as well), and you can proceed to your next action.\\
		Your actions must be based on your previous series of actions, guided by your current position within the overall action path, and carried out in accordance with the characteristics you possess as a social entity from the previous dialogue. Your actions determine your direction of development within the social simulator; please choose your subsequent actions carefully.\\
		Your previous actions are as follows: \{actions\}\\
		The content of the current path node in the action path planning is as follows: \{pathNode\}
		\par\smallskip
		The output requirements are as follows; the output must strictly adhere to JSON format:\\
		1. isSocialize: boolean type; can only be True or False, indicating whether this social entity needs to interact with other social entities.\\
		2. socializeContent: string type. This field describes the purpose of the social entity’s interaction with its target and specific details of the interaction. The description should be as detailed as possible, with a minimum length of approximately 150 characters.\\
		3. actionContent: string type. This field describes the specific details of the social entity’s current action. The description should be as detailed as possible, with a minimum length of approximately 200 characters.\\
		\bottomrule
	\end{tabular}
	\par\vspace{5pt}
	\footnotesize
\end{table}
\clearpage

\section{Algorithm Pseudocode}
This appendix section presents the algorithmic pseudocode for agents interaction processes in social simulation experiment.

\renewcommand{\thealgorithm}{} 
\makeatletter
\makeatother

\renewcommand{\thealgorithm}{} 
\makeatletter
\makeatother

\begin{center} 
	\begin{minipage}{0.9\textwidth} 
		\begin{algorithm}[H]
			\caption{LLM Agent Social Simulation}
			\begin{algorithmic}[1]
				\State \textbf{Initialize:} Set social Agents  $A = \{a_1, a_2, \dots, a_N\}$
				\ForAll{$a_i \in A$ \textbf{parallel}} \Comment{Multi-threaded}
				\State $\text{memory}_i \gets \text{Cold-Start}(\text{dataset}, \text{topic})$
				\State $\text{attributes}_i \gets \text{InitAttributes}()$
				\EndFor
				
				\For{$t = 1 \to T$} \enspace \Comment{Simulation steps}
				\State $\text{BarrierSync}(A)$
				\ForAll{$a_i \in A$ \textbf{parallel}}
				\If{$\text{ShouldInteract}(a_i)$} \enspace \Comment{Interaction decision}
				\State $\text{partners} \gets \text{FindNeighbors}(a_i, \text{radius})$
				\State $\text{available} \gets \emptyset$
				\For{$a_j \in \text{partners}$}
				\If{$\text{RequestConsent}(a_i, a_j)$}
				\State $\text{available} \gets \text{available} \cup \{a_j\}$
				\EndIf
				\EndFor
				\If{$\text{available} \neq \emptyset$}
				\State $\text{ExecuteInteraction}(a_i, \text{available})$
				\EndIf
				\EndIf
				\State $\text{action} \gets \text{GetAction}(a_i)$
				\While{$\neg \text{CheckConstraints}(\text{action})$}
				\State $\text{RemoveAction}(a_i, \text{action})$
				\State $\text{action} \gets \text{GetNewAction}(a_i)$
				\EndWhile
				\State $\text{ExecuteAction}(a_i, \text{action})$
				\State $\text{attributes}_i \gets \text{AdjustAttributes}(\text{attributes}_i, \text{action})$
				\EndFor
				\EndFor
			\end{algorithmic}
		\end{algorithm}
	\end{minipage}
\end{center}
\end{document}